\pdfoutput=1

\documentclass[11pt]{article}

\usepackage[final]{acl}

\usepackage{times}
\usepackage{latexsym}

\usepackage[T1]{fontenc}

\usepackage[utf8]{inputenc}

\usepackage{microtype}

\usepackage{inconsolata}

\usepackage{graphicx}

\usepackage{amsmath}
\usepackage[capitalize,noabbrev,nameinlink]{cleveref}
\usepackage{caption,subcaption}
\usepackage{float}

\usepackage{tipa}

\newif\ifshowcomments
\showcommentsfalse
\newcommand{\isabel}[1]{
    \ifshowcomments
        \textcolor{orange}{[isabel: #1]}
    \else
    \fi}

\title{Using Shapley interactions to understand how models use structure}

\author{%
   \hspace{8mm}Divyansh Singhvi\footnotemark[1]\\
   \hspace{8mm}Independent\\
  \hspace{8mm}\small{\texttt{divyanshsinghvi@gmail.com}}
   \And 
   \hspace{12mm}{Diganta Misra\thanks{equal contribution}}\\
    \hspace{12mm}ELLIS \& MPI-IS Tübingen\\
   \hspace{12mm}\small{\texttt{diganta.misra@tue.ellis.eu}}
   \And
   \hspace{16mm}Andrej Erkelens\footnotemark[1]\\
   \hspace{16mm}Independent\\
   \hspace{16mm}\small{\texttt{andrej.erkelens@gmail.com}}
   \And
   Raghav Jain\footnotemark[1]\\  
   UC San Diego \\
  \small{\texttt{r6jain@ucsd.edu}}
  \AND
  Isabel Papadimitriou\thanks{equal contribution} \\
  Kempner Institute, \\
  Harvard University \\
  \small{\texttt{isabelpapadimitriou@fas.harvard.edu}} \\
  \And
  Naomi Saphra\footnotemark[2]\\
   Kempner Institute, \\
  Harvard University \\
  \small{\texttt{nsaphra@fas.harvard.edu}} \\
}

\begin{document}
\maketitle

\begin{abstract}
Language is an intricately structured system, and a key goal of NLP interpretability is to provide methodological insights for understanding how language models represent this structure internally. In this paper, we use Shapley Taylor interaction indices (STII) in order to examine how language and speech models internally relate and structure their inputs. Pairwise Shapley interactions measure how much two inputs work \textit{together} to influence model outputs \textit{beyond} if we linearly added their independent influences, providing a view into how models encode structural interactions between inputs. We relate the interaction patterns in models to three underlying linguistic structures: syntactic structure, non-compositional semantics, and phonetic coarticulation. We find that autoregressive text models encode interactions that correlate with the syntactic proximity of inputs, and that both autoregressive and masked models encode nonlinear interactions in idiomatic phrases with non-compositional semantics. Our speech results show that inputs are more entangled for pairs where a neighboring consonant is likely to influence a vowel or approximant, showing that models encode the phonetic interaction needed for extracting discrete phonemic representations.

\end{abstract}

\section{Introduction}

How do language model features work \textit{together} to influence prediction results? Do the internals of language models reflect the complex structure of language in how they combine features? Feature attribution---measuring how different model features (like inputs or neurons) influence output decisions in isolation---is a key method of understanding and interpreting neural models. One common approach to feature attribution is adapted from game theory scenarios, and treats features like agents in a cooperative game, attributing credit for the outcome to each feature \citep{DBLP:journals/corr/LundbergL17}. This credit value, or \textbf{Shapley value} \citep{shapley1952value}, quantifies the additive effect of each feature on the output, assuming that features act in a linearly independent manner on the output. The linearity assumption is not accurate for most deep learning scenarios: neural networks are non-linear, and features interact in complex ways inside model representations to influence output predictions. 

What interactions between features do we miss when we assume this linear independence? To address this question, researchers have proposed methods to calculate how much information we lose when assuming additivity \citep{kumar2021residuals}, and \textbf{Shapley interactions}, accounting for how features have influence in pairs or groups on top of how they act independently \citep{agarwal2019interaction}.

\textbf{In this paper, we investigate how Shapley interactions can enhance our understanding of the internal processes of language models.} We ground our investigation in structural features that we know about the input data (like syntactic structure), and ask: what do Shapley interactions reveal about how the model uses the dependency structure in language? By relating Shapley interactions to structural linguistic features, we showcase how different models use (or don't use) linguistic structural features in their internal representations.  We run experiments on autoregressive and masked text models, as well as on automatic speech recognition models, and report the following findings:

\begin{itemize}
    \item Autoregressive models (but not masked models) show a strong correlation between Shapley interaction and the syntactic proximity of features. This result indicates that syntactic structure is encoded in non-linear interactions between model features (\Cref{sec:syntax}).
    
    \item Both autoregressive and masked models exhibit stronger interactions between pairs of tokens in \textit{multiword expressions} (MWEs) that have idiomatic non-compositional meaning (expressions like \textit{to kick the bucket}, meaning \textit{to die})  (\Cref{sec:mwes}).

    \item In speech models, Shapley interactions are stronger between consonants and vowels than between pairs of consonants (\Cref{sec:vowels}). This result accords with the known phenomenon of \textit{coarticulation}: the acoustics of vowels are often shaped by the surrounding consonants, while consonants can be understood in isolation \citep{rakerd1984vowels}. This finding also extends to more sonorant vowel-like consonants, which interact more with surrounding consonants than those produced with the vocal tract more closed (\Cref{sec:mannerofarticulation}).
\end{itemize}

The non-linearities and interactions in model internals are a vital missing piece of the wider language model interpretability inquiry.
Our work showcases Shapley interactions as a powerful interpretability method for examining how language models organize their representations to reflect latent input structure.

\section{Background and related work}

\begin{figure*}[!ht]
    \centering
    \begin{subfigure}{0.49\textwidth}
      \centering
      \includegraphics[width=\linewidth]{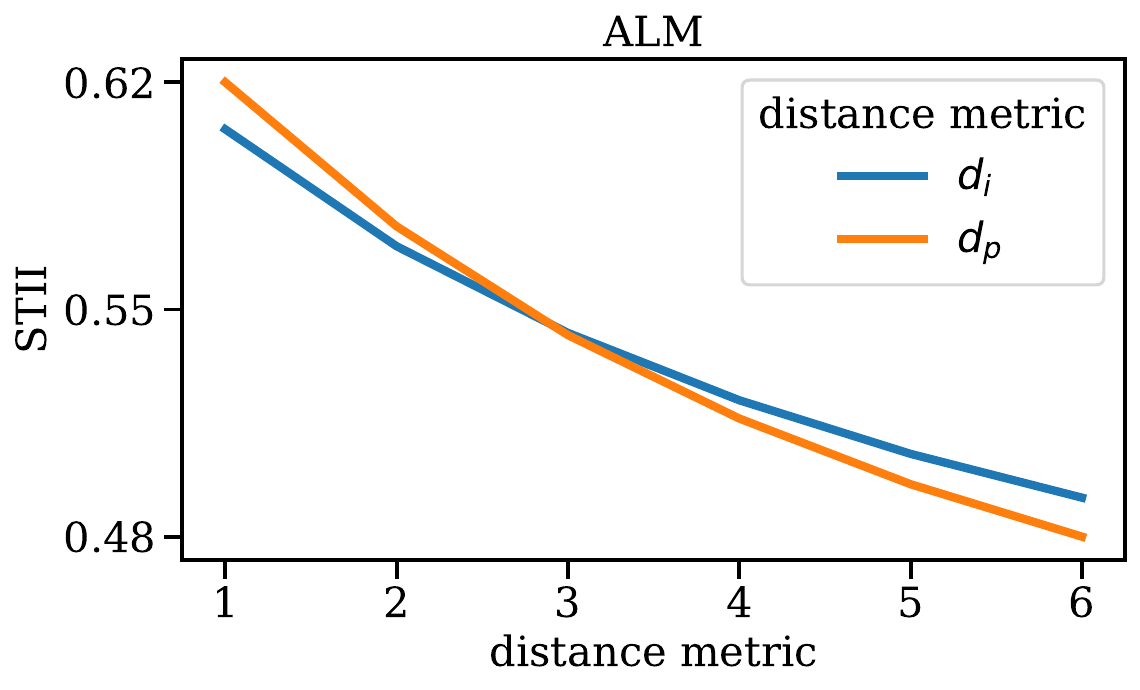}
      \label{fig:llm_avg}
    \end{subfigure}
    \centering
    \hfill
    \begin{subfigure}{0.49\textwidth}
      \centering
      \includegraphics[width=\linewidth]{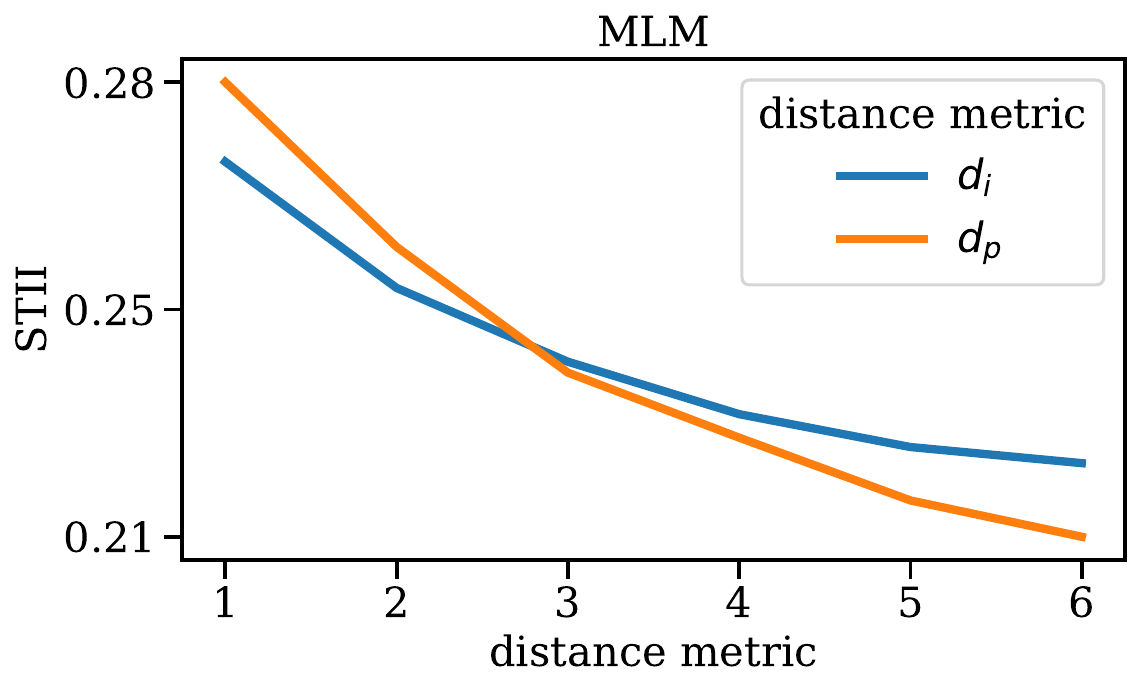}
      \label{fig:mlm_avg}
    \end{subfigure}
    \caption{\textbf{The effect of interacting pair distance and prediction distance on feature interactions.} Our results for the experiments relating Shapley Interactions with a token's position in the sequence. We find that, for both autoregressive models (left) and masked models (right), STII decreases monotonically with distance. This holds both when we are measuring distance as the distance between the two elements in the interacting pair ($d_i$, blue line) and when we are measuring distance between the interacting pair and the token that the model is predicting ($d_p$, orange line). Our results indicate that models treat tokens that are far away from each other more like an unentangled bag-of-words, and that they treat pairs of tokens that are far away from the token being predicted as unentangled, no matter the distance between them. 
    \isabel{I would maybe make one of the lines dashed (this looks like a colorblind-friendly pallette, but it's still good to do)} }
    \label{fig:avg}
\end{figure*}

\subsection{Shapley Interactions}

Shapley values are used to attribute decisions to specific features in predictive models.
The Shapley value of a set of features $A$ is obtained by computing the difference in a model's output when $A$ is included versus when it is excluded. If we take the set of all features $N$ and remove $A$, we want to see how much value $A$ adds to every possible context subset $S \subseteq N \backslash A $. In our case, the value function $v$ is the logit output of the model. The Shapley value is the weighted average of this marginal contribution over all $S$:

\begin{equation}
\begin{split}
    &\phi(A) = \sum_{S\subseteq N \backslash A} w_S \left(v(S\space\cup A) - v(S)\right) 
\end{split}
\end{equation}

where the weight $w_S$ for each subset is the number of possible subsets $S$ of the same size:
\begin{equation}
    w_S =  {{|N| - |A|} \choose |S|}
\end{equation}

If the features are linearly additive without interactions, note that $\phi(\emptyset) \approx \sum_{i \in N} v (\{i\})$.  However, in scenarios where features are dependent and their composition is non-linear, Shapley values do not account for interacting effects between sets. Various methods to quantify and use these interactions have been proposed \citep{owen1972multilinear,grabishroubens,fumagalli2023shapiq,tsai2022faithshap,kumar2021residuals}. 

For simplicity, we consider only pairwise interactions between two feature sets $A$ and $B$. To calculate pairwise Shapley interactions, we rely on the \textbf{Shapley Taylor interaction index} (STII)  \citep{agarwal2019interaction} to calculate second-order interactions using the discrete second-order derivative. Since our features are vectors, we calculate the scalar Shapley interaction value for each dimension individually, and use the norm of this vector as an interaction metric. 
 Similar to \citet{saphra-lopez-2020-lstms}, we scale the result by the norm of the entire sequence with no feature ablations.
\begin{equation}
\textrm{STII}_{A,B} = \frac{\lVert \phi(\emptyset) - \phi(A) - \phi(B) + \phi(A, B) \rVert_2}{\lVert \phi(\emptyset) \rVert_2}
\label{eq:interaction}
\end{equation}

Calculating the Shapley values for each coalition requires iterating over the powerset of $N$, requiring $O(2^{|N|})$ calculations. In high-dimensional input spaces, the exact calculation of Shapley residuals is therefore prohibitively expensive. We compute the Shapley value using the method of \citet{mitchell2022sampling}, which approximates the weighted average of subset credits with Monte Carlo Permutation Sampling \citep{Castro2009PolynomialCO}.

\subsection{Structure in language models}

There is a huge and varied literature aimed at understanding how language models use and represent the structure in their  input. Approaches include examining if the output probabilities of language models reflect structural rules \citep[][\textit{inter alia}]{DBLP:journals/corr/abs-1805-12471, hu2024align, gauthier2020syntaxgym}, as well as looking inside model representations. For the latter approaches, while many linguistic structural elements can be linearly extracted from the representations of text and speech models \citep[][\textit{inter alia}]{hewitt-manning-2019-structural, belinkov2021probing, pasad2024selfsupervised, chrupala-etal-2020-analyzing, park2023linear}, and attribution methods can relate the linear importance of different features in text and speech models \citep[][\textit{inter alia}]{Markert_2021, ethayarajh2021attention, yeh2020completeness, kokalj2021bert}, the fact remains that neural models have complex nonlinearities in their internal processing. 

How can we analyze the ways in which nonlinear interactions play out in model internals, and what they encode? Multiple papers have analyzed the difficulties of knowing what we can extract when using nonlinear probing methods \citep{voita2020information,pimentel2021bayesian, hewitt2021conditional}, and others have proposed searching for causal effects which can be generally agnostic to whether the processing is linear \citep{geiger2021causal, arora2024causalgym}. 
Shapley interactions let us directly link features of the input to different \textit{extents} of nonlinear processing. Prior work showing the utility of Shapley interactions in analyzing NLP models has focused on older architectures like LSTMs \citep{saphra-lopez-2020-lstms,singh2018hierarchical} and on models fine-tuned for simple text classification tasks \citep{jumelet-zuidema-2023-feature, chen2020generating}.
Our work builds on and generalizes these results by relating Shapley interactions to diverse forms of linguistic structure (syntactic, semantic, and phonetic) on models trained on domain-general language tasks (generation for text, and ASR for speech)

\section{Text models: Interactions between tokens}

Our first experiments are on language models, measuring how known associations between tokens correlate with Shapley-based measures of feature interaction. We consider the influence of token position, idiomatic phrases, and syntax. We find that masked LMs and Autoregressive LMs differ in their interaction structure, especially in how they respond to syntax.

\paragraph{Models and Datasets}
We run all of our experiments on two models: the autoregressive model GPT-2 \citep{radford2019language} and the masked language model BERT-base-uncased \citep{DBLP:journals/corr/abs-1810-04805}. Each input sentence is unpadded and truncated to 20 tokens, and we apply softmax to the logit outputs to ensure that interactions across different examples are comparable. 

All English language modeling experiments use \texttt{wikitext-2-raw-v1} \citep{merity2016pointer} tokenized and dependency parsed (for syntax experiments) with spaCy \citep{spacy2}. We resolve incompatibilities between the spaCy tokenizer and the model-specific tokenizers by assigning overlapping tokens a syntactic distance of zero. For the multiword expression experiments, we use the AMALGrAM supersense tagger \citep{schneider-etal-2014-discriminative}, which identifies both strong and weak \citep{21b3891a2f3841febb0ce3ebd3cc4c4a} MWEs.

  \begin{figure*}[!ht]
    \centering
    \begin{subfigure}{0.49\textwidth}
      \centering
      \includegraphics[width=\linewidth]{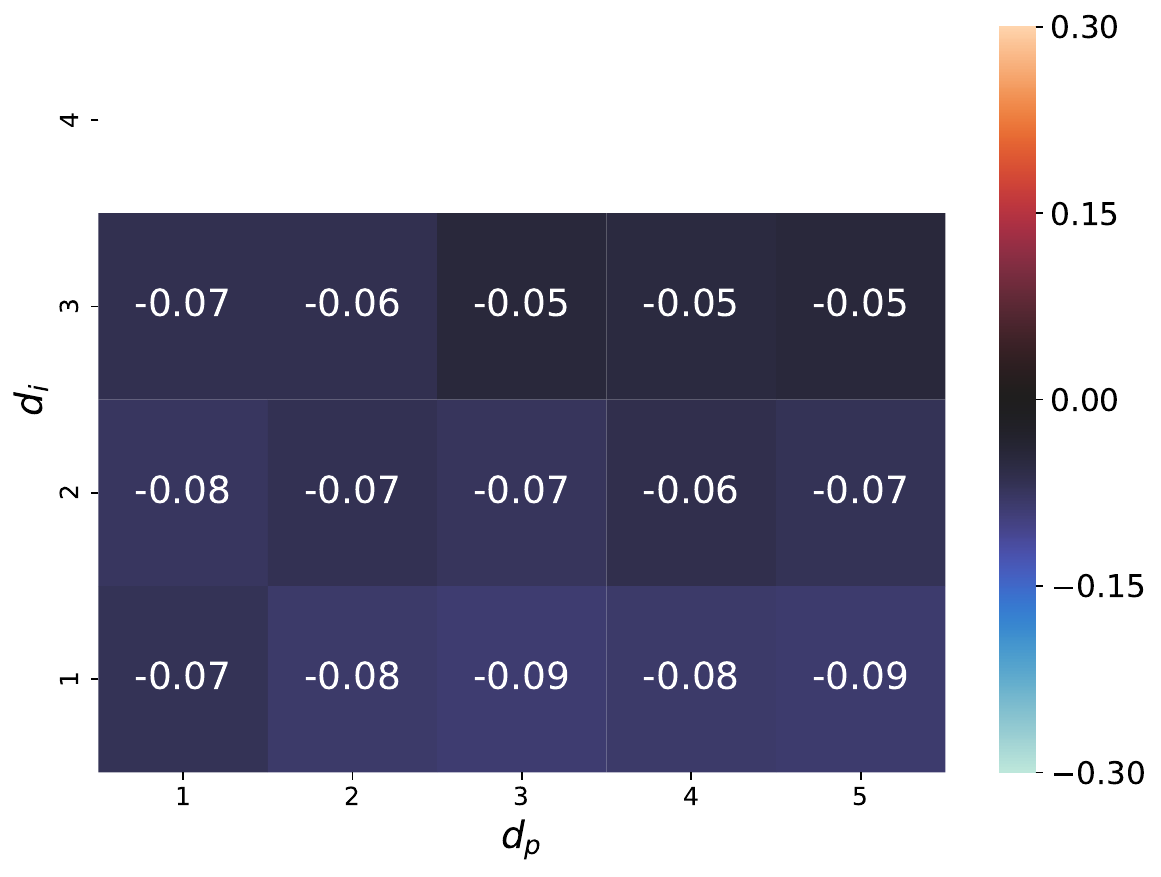}
      \caption{Autoregressive language model experiments}
      \label{fig:llm_heatmap}
    \end{subfigure}
    \centering
    \hfill
    \begin{subfigure}{0.47\textwidth}
      \centering
      \includegraphics[width=\linewidth]{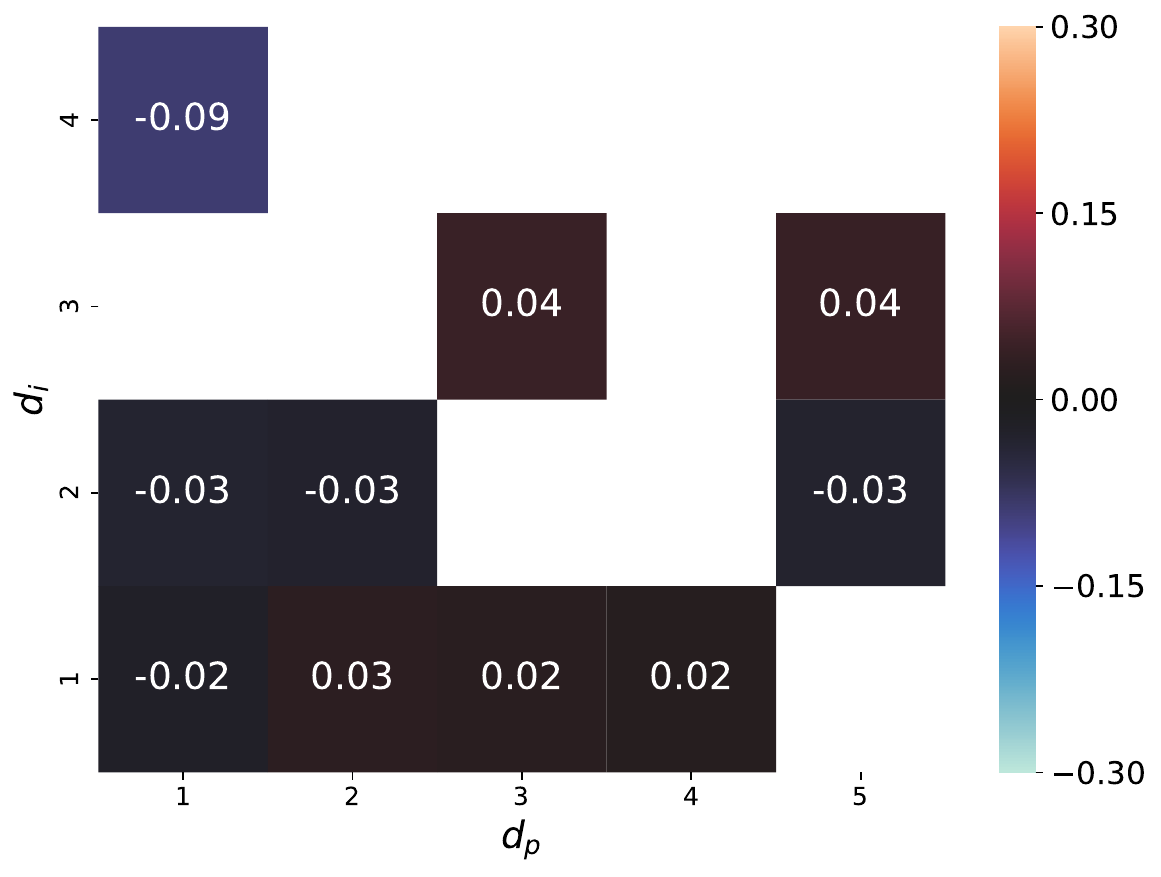}
          \caption{Masked language model experiments}
      \label{fig:mlm_heatmap}
    \end{subfigure}
    \caption{\textbf{Syntactically close token pairs interact more than other pairs with the same positional distance.} The results of our syntactic distance experiments (\Cref{sec:syntax}): how does syntactic distance correlate with STII, controlling for the effect of position? A negative correlation means that tokens closer in the parse tree (low syntactic distance) are more heavily entangled (high STII). Autoregressive models show a consistently negative correlation in all significant cells, meaning that syntax is encoded in Shapley interactions. We stratify our results by the two positional distance metrics in \Cref{sec:position}, so that we can calculate the effect of syntactic distance, marginalizing out the effect of positional distance. Each cell displays a correlation between syntactic distance and STII for a given interacting pair distance and prediction distance. We only provide results for cells where there exists at least one direct syntactic modifier pair separated by the positional distance $d_i$ and the Spearman correlation given at that cell is statistically significance ($p < 0.05$). For our correlation calculation, we only include a syntactic distance if there are at least 50 data points with that syntactic distance in our data set. \isabel{I think that this figure is a bit hard to parse at first. Space-allowing, it might be nice to have another figure for this section that shows like, individual points having a negative correlation or something. Maybe even like, a magnifying glass on one of these squares? Also, this heat map pallette is a bit unintuitive to me. I would think that the darker blues are ``more'' for the left figure, but instead they are less.} }
    \label{fig:heatmap}
  \end{figure*}

\subsection{Baseline: the effect of position}
\label{sec:position}
One potential factor influencing interactions between tokens is the positional distance between tokens
Let's say that we are calculating the interaction between two tokens, $x_{t_1}$ and $x_{t_2}$ at positions $t_1$ and $t_2$. The token that the model is trying to predict (i.e. the next token in autoregressive models, and the masked token in masked models) is at position $t_\text{target}$. There are two relevant positional distances that are likely to influence interaction.

Firstly, the \textbf{interacting pair distance}, $d_i$, is the distance between the two tokens: 

\begin{align}
    d_i(x_{t_1}, x_{t_2}, x_{t_{\textrm{target}}}) &= t_2 - t_1 \label{eq:pair_distance}
\end{align}

Secondly, the \textbf{prediction distance}, $d_p$, is the distance between the pair of tokens that we are calculating the interaction of, and the target token that the model is trying to predict:
\begin{align}
    d_p(x_{t_1}, x_{t_2}, x_{t_{\textrm{target}}}) &= \min_{t \in \{t_1, t_2\}} \left| t_{\textrm{target}} - t \right| \label{eq:prediction_distance}
\end{align}

For our position baseline experiments, we test how both interacting pair distance and prediction distance influence the STII between the two tokens $x_{t_1}$ and $x_{t_2}$.

\paragraph{Results}

Our results are presented in \Cref{fig:avg}, confirming that distance has an effect on STII in both autoregressive and masked models. This holds whether we are measuring distance as distance between the interacting pair (interacting pair distance $d_i$) or distance between the last token in that pair and the target prediction token (prediction distance $d_p$). The dramatic decline of STII with increased prediction distance implies that when these models predict tokens, they treat the more distant context as a bag of words rather than as complex syntactic relations \citep{khandelwal-etal-2018-sharp}. We also see that closer tokens interact more strongly with each other. 

For the rest of our experiments, we will stratify samples by both $d_i$ and $d_p$, so that we can measure the effects of linguistic structure \textit{beyond} these  position effects that we demonstrate here.

  \begin{figure*}[ht!]
    \centering
    \begin{subfigure}{\textwidth}
      \centering
      \includegraphics[width=\linewidth]{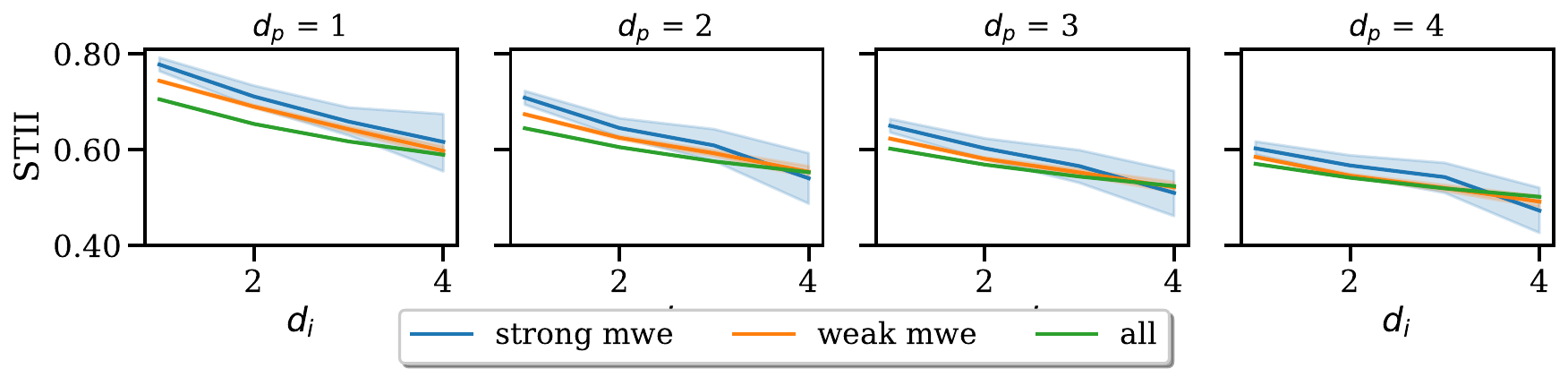}
      \caption{Autoregressive LM experiments}
      \label{fig:llm_experiment_comparison}
    \end{subfigure} \\
    \begin{subfigure}{\textwidth}
      \centering
      \includegraphics[width=\linewidth]{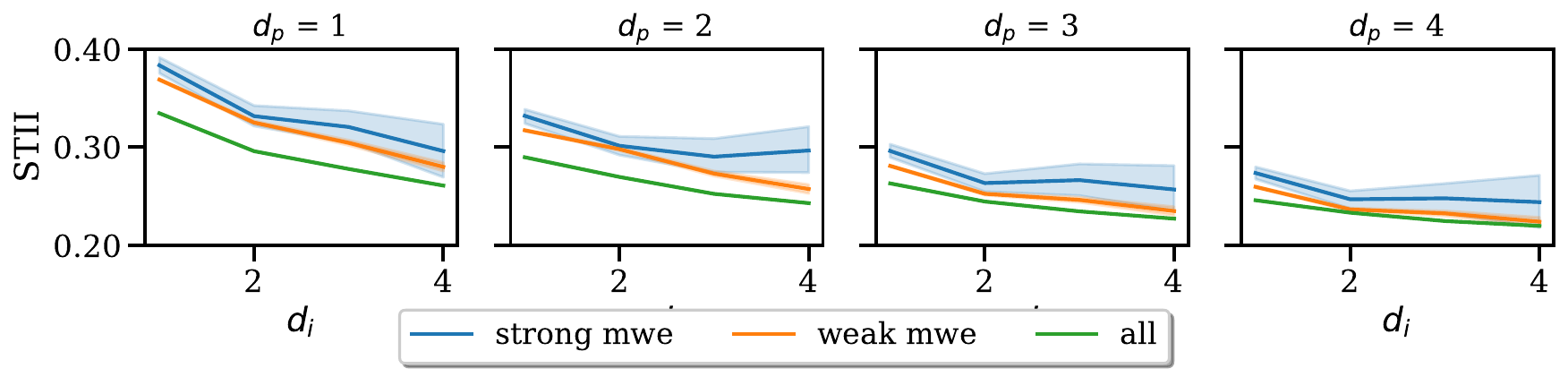}
      \caption{Masked LM experiments}
      \label{fig:mlm_experiment_comparison}
    \end{subfigure}
    \caption{\textbf{Tokens interact more if they occur in the same multiword expression:} Shapley interactions are higher for tokens in multiword expressions than tokens that are not. 
    The results are controlled for prediction distance $d_p$ (different facets) and interacting pair distance $d_i$ (x-axis). Within each facet for each x-axis value, we can see that the STIIs for tokens in Strong MWEs (blue) and Weak MWEs (orange) are significantly higher than the average over all pairs (green).}
    \label{fig:mwes_experiment}
  \end{figure*}

\subsection{Syntactic structure}
\label{sec:syntax}

Syntactic structure can also influence an LM's predictions. If a model composed distant syntactic relations in a linear way, it would treat the wider context as though it were a bag of words. By instead exhibiting strong interactions between syntactically close tokens, the model would closely entangle the meaning of a modifier with its head. We measure \textbf{syntactic distance} by the number of dependency edges traversed to connect a pair of tokens, a metric encoded by projected representations in both masked \citep{hewitt-manning-2019-structural} and autoregressive \citep{murty_characterizing_2022} models. We verify the role of modifier connections by the Spearman correlation between syntactic distance and STII, stratified by interacting pair distance and prediction distance. %

\paragraph{Results}

\Cref{fig:heatmap} shows correlation between syntactic distance and STII.  Our analysis reveals that, for autoregressive language models, all statistically significant correlations are negative. In contrast, non-autoregressive language models exhibit both positive and negative correlations. This finding aligns with \citet{saphra-lopez-2020-lstms}'s research on LSTMs showing that syntax is handled more consistently in autoregressive models, and with \citet{ahuja2024learning}, who in a different setting show that autoregressive models are more predisposed to syntax-style generalizations. 

The inconsistencies observed in non-autoregressive models may stem from their handling of positional proximity in less intuitive ways, complicating the relationship between syntactic and linear distance. The interaction between these two dimensions may be more difficult to manage in masked models, leading to the varied correlation outcomes.

This finding suggests that we can interpret feature interaction as a distinctly syntactic alternative to the inherent distance encoding found in autoregressive architectures \citep{haviv2022transformer}. In these models, the degree of interaction is learned to prioritize syntactic relationships rather than depending solely on positional information within the language modeling objective. This highlights a fundamental difference in how these models integrate syntactic structure and distance.

\subsection{Multiword expressions}
\label{sec:mwes}

While semantics is often treated as compositional (the meaning of a sentence can be composed by rules, following the syntax and the meaning of each individual word), language is also characterized by non-compositional, or idiomatic, phrases. These are groups of words whose meaning can only be derived when looking at the entire group rather than the individual words. These word groups, known as \textbf{multiword expressions} (MWEs), include idioms like \textit{break a leg}, where the isolated meaning of each of the component words \textit{break}, \textit{a}, and \textit{leg} fail to compose the meaning of the entire expression. Higher interaction values for the tokens in the idiom would indicate a less compositional treatment of the whole phrase.

We use the MWE tagger from \citet{schneider-etal-2014-discriminative}, which also distinguishes strong and weak MWEs. As explained by \citet{schneider2014comprehensive}, strong MWEs are non-compositional idiomatic phrases, in which the meaning of each word is changed substantially by its context. Weak MWEs, meanwhile, are sequences with distributional properties similar to strong MWEs, but where the meaning of each word is maintained. In other words, \textit{near miss} would be a weak MWE---it means an event that was \textit{nearly} a \textit{miss}---but \textit{close call} would be a strong MWE, as the meanings of \textit{close} and of \textit{call} are both tied to the particular idiom.

In these experiments, we compare interactions between arbitrary pairs of tokens to interactions between tokens contained within an MWE. The extreme case where there is no Shapley residual would imply perfect compositionality---after all, linear addition is compositional---so our hypothesis is that MWEs have a larger than average residual.

    \begin{figure}[ht]
      \centering
      \includegraphics[width=\linewidth]{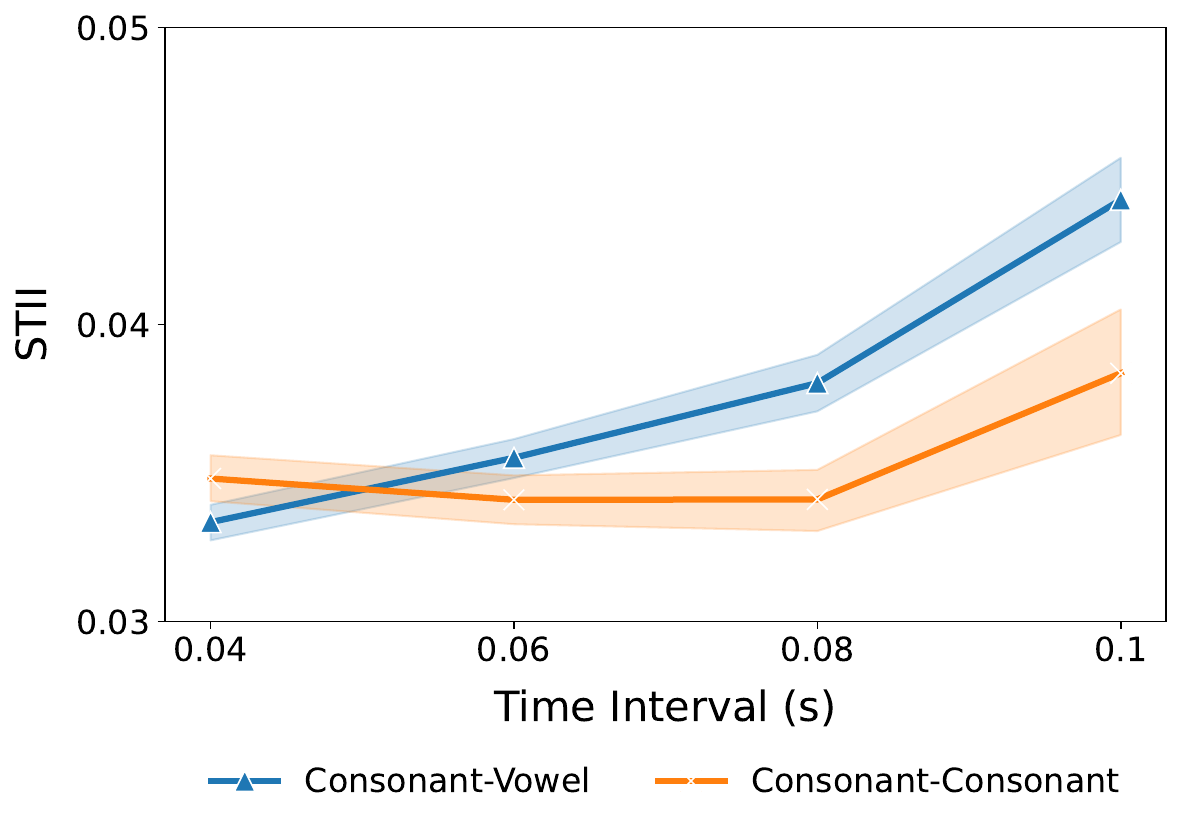}
          \caption{\textbf{Vowel-consonant interactions are higher than consonant-consonant interactions.} Average STII between pairs of adjacent acoustic input features, adjusting the interval range around the phone boundary. Confidence intervals are provided by bootstrap.}
        \label{fig:vowel_consonant}
    \end{figure}

\paragraph{Results}

\Cref{fig:mwes_experiment} compares the STII between tokens that belong to the same MWE to the average STII between all tokens, stratified by interacting pair distance  $d_i$ and prediction distance  $d_p$. 
For both the autoregressive models (\Cref{fig:llm_experiment_comparison}) and masked models (\Cref{fig:mlm_experiment_comparison}), STII is higher when the interacting pair is in a MWE: the blue and orange MWE lines are overall higher in STII than the green baseline.  The effect is consistent across positional distances and more pronounced when predicting nearby tokens. %

Importantly, the strong MWEs also have higher interaction than weak MWEs. This result confirms that the nonlinear interactions are driven by semantic dependencies between tokens, and not only by distributional properties, like high co-occurrence, which are shared by both weak and strong MWEs.

    \begin{figure*}
      \centering
      \includegraphics[width=\linewidth]{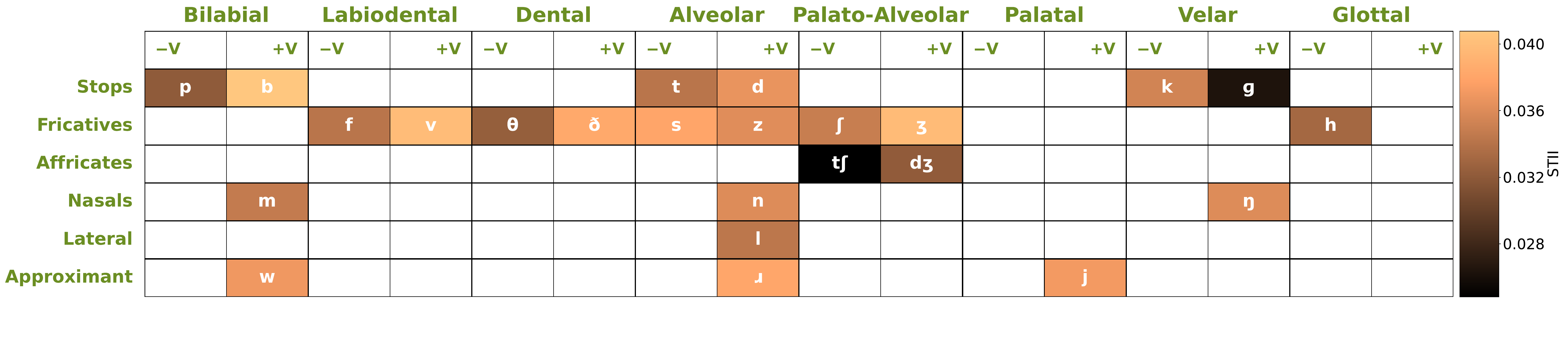}
          \caption{\textbf{In consonants, feature interactions are determined by sonority.} Consonant chart with a heat map indicating average interaction with acoustic features from adjacent phonemes (samples from 0.1s around the phoneme boundary). Columns indicate the place of articulation while rows indicate the manner of articulation. Only interactions for acoustic features within 0.1s range around the phoneme boundary are considered. Consonants with more vowel-like articulations (lower down in the chart) tend to have higher interactions with surrounding phonemes. Consonant voicing is denoted by \texttt{-V} and \texttt{+V}, which represent Unvoiced and Voiced consonants respectively.
              }
        \label{fig:consonants}
    \end{figure*}

\section{Speech models: Interactions between phones}

Do speech models represent phonetic interactions? Consonants influence the realization of vowels, and in order to be able to separate vowels into a consistent discrete system a listener has to take these interactions into account \cite{rakerd1984vowels, rosner1994vowel}. Vowels are produced in a continuous space, without clear boundaries that delineate which vowel a specific vocal tract positioning refers to (for example, a speaker can glide on the continuum between [i] and [e], but there is no clear analog of a continuum between [p] and [k]). The realization of vowels is influenced by the consonants that surround them. 
Despite the continuous nature of vowel phonetics, listeners perceive vowels as belonging to a few discrete classes of vowel phonemes. To derive this discrete phonological representation, a listener --- or predictive speech model --- would need to represent the structure of consonant-vowel interaction to correctly identify a vowel phoneme from its phone. We use Shapley interactions to demonstrate the elevated importance of nearby phonemes for processing vowels, as compared to consonants.

Since the inputs of speech models are not cleanly tokenized into phones, and the transition between phones is continuous and without a well-defined boundary, we measure interaction by taking the average pairwise interaction within a time interval that includes a transition.
For a given interval length, we measure STII between all temporally consecutive features $p_{t_1}$ and $p_{t_2}$ when predicting the immediate next sound $p_{t_3}$. Formally, the interaction $N$ between different phonemes over a temporal interval within range $\delta$ of the approximated phone boundary time $t_b$ is:
\begin{equation}
\bar{r}_{\delta} = \sum_{t_1 = t_b - \delta}^{t_b + \delta} \text{STII}_{p_{t_1}, p_{t_2}}
\end{equation}
Note, however, that in the case where no acoustic feature is sampled at exactly $t_b - \delta$, we instead start the summation with $t_1$ at the earliest timestamp such that $t_1 \geq t_b - \delta$. Since all interaction pairs are consecutive, the confounder of positional distance is automatically removed for these experiments.

\paragraph{Models and Datasets} Our experiments are run on the Wav2Vec 2.0 model wav2vec2-base-960h \citep{baevski2020wav2vec}, which is trained on 960 hours of English audio to predict the next sound in a recording. When computing Shapley values, ablated acoustic features are replaced with silence. 

For all experiments, we use the Common Voice dataset \citep{commonvoice:2020} of English language voice recordings, which are contributed by volunteers around the world and comprise 92 hours of recorded speech. This compilation is characterized by its rich diversity, featuring a total of 1,570 unique voices. Specifically, we use a set of 198 audio files totaling approximately 20 minutes of audio. We preprocess the dataset by aligning the audio recordings with their corresponding phonemes using Montreal Forced Aligner (MFA)~\citep{mcauliffe17_interspeech}, which uses acoustic models to map the audio recordings to their corresponding phonemes. We preprocess all audio files to a WAV and standard sampling rate (16kHz) and then use the US ARPA acoustic model~\citep{mfa_english_us_arpa_acoustic_2024} from MFA to detect and align phonemes within the speech to their corresponding timeframes in the recordings, marking the start and end of each phoneme. It is important to note, as a caveat to the following results, that identifying the exact duration of a phoneme is not only challenging but undefined in practice, as the vocal tract is in a state of continuous transition between phonemes throughout an utterance.

\subsection{Interactions between consonants and vowels}
\label{sec:vowels}

Vowels are formed with an open vocal tract that produces no turbulent airflow, with the specific position of each part of that anatomy
largely determined by the surrounding consonants. Therefore, it is harder to map vowel sounds in isolation to their corresponding discrete phoneme than it is to map consonants \citep{rakerd1984vowels}. In \Cref{fig:vowel_consonant}, we compare the interactions over consonant-vowel boundaries and consonant-consonant boundaries, and find that interactions are significantly higher in the consonant-vowel case. This implies that the model is taking this entanglement into account, which is necessary for reaching a discrete phonological analysis of the input similar to human phonological perception.

\subsection{The effect of consonant manner of articulation}
\label{sec:mannerofarticulation}

Not all consonants are equally stable in their capacity to be interpreted in isolation. In describing consonants, the \textit{manner of articulation} refers to a hierarchy of vocal tract occlusion, ranging from the stops (consonants like [p], formed by briefly blocking all air through the vocal tract) to the approximants (consonants like [j] as in ``universe'', that produce only slightly more turbulent airflow than vowels).  Therefore, some consonants in practice behave more like vowels, and we expect them to exhibit more nonlinear interactions across phoneme boundaries, as vowels do.

Our hypothesis is largely confirmed in \Cref{fig:consonants}, modeled on a International Phonetic Alphabet consonant chart where row indicates the manner of articulation.
Although the pattern is not perfect, the figure shows high cross-phoneme STII for more sonorant consonants on the lower rows, which are articulated like vowels with a more open oral cavity. Furthermore, voiced consonants tend to have higher STII than their unvoiced counterparts, as expected from the turbulent vibrations they command. The voicing relationship is visible in the pairs [p]/[b], [t]/[d], [f]/[v], [\textipa{T}]/[ð], [\textipa{S}]/[\textipa{Z}], [\textipa{tS}]/[\textipa{dZ}], but not  in [k]/[g] (where [g] is an outlier with very high STII) or in [s]/[z] (which are close in value). %

\section{Future Work}

Our primary objective in this work has been to showcase the versatility of Shapley interactions in showing the ways that language models encode linguistic structure. Understanding structural representation, and especially how this can be nonlinear, is a long-standing problem and inquiry in NLP interpretability. This work suggests a number of open questions and follow-up problems, in addition to having the potential to be applied as is to different types of annotated linguistic structure.

Speech has multiple layers of structure, as it comprises both an acoustic signal and the language structure underlying the utterance. Our investigation of feature interactions is limited to the phonetic level, but future work may find the degree to which these multiple layers of linguistic structure affect nonlinear feature interactions. Do these speech models exhibit similar interaction patterns to the autoregressive language models we also analyze? Speech, often neglected in interpretability research, is ripe with open problems.

While we compare the behavior of the models trained on the masked and autoregressive objectives, we do not compare any models that are trained on the same objective with different architectures. The inductive bias and function of a given architecture are matters of great interest to many researchers in machine learning, and we believe that measuring nonlinear interactions can provide many insights into how specific models are similar and different.

This work focuses on pairwise interactions, and so has not taken full advantage of the versatility of Shapley residuals as a tool. Higher order Shapley interactions \citep{sundararajan2020shapley} provide a method of hierarchical clustering on features and introduce yet more nuance into approximations of linear and nonlinear behavior in neural networks. We also do not consider interactions of internal model features. We suggest that future work in the area should incorporate knowledge about the underlying semantics of the input as well as the model architecture. 

Finally, and most crucially, we believe that followup work in this area should be interdisciplinary. Speech, language, image processing, and other areas that can benefit from interpretability are all well-studied, with decades or even centuries of scientific research. By collaborating with specialists in these data domains, we can potentially contribute not only to the understanding of artificial models, but also to the understanding of the natural phenomena in question. Interpretability is an important new area in the emerging field of AI for scientific understanding and discovery, and we encourage others to start future work by finding domain experts to choose questions worth asking.

\section{Conclusions}

In accordance with The Bitter Lesson \citep{sutton2019bitter}, researchers and engineers typically apply machine learning methods generically, incorporating as little explicit data structure as possible. However, The Bitter Lesson does not apply to \textit{interpretability}. Instead, meaningful interpretations of representational and mechanistic structures at scale should be informed by the underlying structure of data. Our results show how to use constituents, phones, and object boundaries to build a scientific understanding that goes beyond intuitions about n-grams, acoustic features, and pixels.

These results have spanned modality and task.
By measuring feature interaction in language models, we present a novel way of describing how the hierarchy of syntactic structure and the encoding of non-compositional semantics both function in model internal representations.  
In speech prediction models, we show that consecutive acoustic features near a phone transition have more nonlinear interactions if the transition is between a consonant and vowel, rather than between two consonants. We also see that in this sense, sonorant consonants behave more like vowels. 

These studies do not focus on individual data samples, but on patterns in the structure underlying the data. Understanding these general patterns requires greater domain expertise than is often required for sample-level interpretability research. We hope to inspire future interdisciplinary work with phonology, syntax, visual perception, and other sciences that characterize corpus-wide structural phenomena.

\section*{Limitations}

The work in this paper shows correlations between pairwise Shapley interactions and structural relationships between two inputs. Both the pairwise aspect, and the fact that we only do correlational analyses, are limitations. There are two ways to expand the analysis to make it more descriptive and informative about the internal processing of models. Firstly, we could look beyond pairwise interactions, creating a hierarchy of interaction: single feature, pairwise, groups of three features, etc. This hierarchy of interaction could be related to more subtle and hierarchical features. While currently we're limited to pairwise features like syntactic proximity, we could more fully analyze complex tree structure if we had a hierarchy of interaction effects. The second way in which this analysis could be made stronger would be to go beyond looking at correlations, and investigate the causal predictive power of Shapley interactions, and the ways in which they change the structural processing and effects of language models. 

The analyses in this paper are not on model sizes close to the order of magnitude of state-of-the-art production models, meaning that the specifics of our results might not be relevant to the models that are having the most effect on the world at the moment. Our paper is meant to showcase the applicability of STIIs to relating model internals to structure in the input, and like all interpretability methods introduced on smaller models, we hope that the viewpoint and methodologies of this paper can be applied to larger models in the future as the field and our understanding develops. 

We do not forsee significant risks from the application and development of this interpretability methodology. 

\section*{Acknowledgements}
This work was supported by Hyundai Motor Company (under the project Uncertainty in Neural Sequence Modeling) and the Samsung Advanced Institute of Technology (under the project Next Generation Deep Learning: From Pattern Recognition to AI). This work has been made possible in part by a gift from the Chan Zuckerberg Initiative Foundation to establish the Kempner Institute for the Study of Natural and Artificial Intelligence. 
    
This collaboration was enabled by the ML Collective (MLC). The authors thank the International Max Planck Research School for Intelligent Systems (IMPRS-IS) for supporting Diganta Misra. This work was partially enabled by compute resources provided by Mila\footnote{\url{https://mila.quebec}} and was funded by the Max Planck \& Amazon Science Hub. 

\bibliography{custom}

\end{document}